\documentclass{article}
\usepackage{spconf,amsmath,amssymb,graphicx}
\usepackage{booktabs}
\usepackage{url}

\title{Applied Federated Learning: \\
Improving Google Keyboard Query Suggestions}

\name{Timothy Yang*, Galen Andrew*, Hubert Eichner*}
\secondname{Haicheng Sun, Wei Li, Nicholas Kong, Daniel Ramage, Fran{\c{c}}oise
Beaufays}

\address{
  Google LLC,\\
  Mountain View, CA, U.S.A.\\
  \texttt{\{timyang, galenandrew, huberte}\\
  \texttt{haicsun, liweithu, kongn, dramage, fsb\}@google.com}
}

\begin{document}
\maketitle

\begin{abstract}

Federated learning is a distributed form of machine learning where both the
training data and model training are decentralized. In this paper, we use
federated learning in a commercial, global-scale setting to train, evaluate and
deploy a model to improve virtual keyboard search suggestion quality without
direct access to the underlying user data. We describe our observations in
federated training, compare metrics to live deployments, and present resulting
quality increases. In whole, we demonstrate how federated learning can be
applied end-to-end to both improve user experiences and enhance user privacy.

\end{abstract}

\section{Introduction}
\label{sec:introduction}

The introduction of \textit{Federated Learning} (FL)~\cite{fedlearn, flstrats,
fedopt} enables a new paradigm of machine learning where both the training data
and most of the computation involved in model training are decentralized. In
contrast to traditional server-side training where user data is aggregated on
centralized servers for training, FL instead trains models on end user devices
while aggregating only ephemeral parameter updates on a centralized server. This
is particularly advantageous for environments where privacy is paramount.

The Google Keyboard (Gboard) is a virtual keyboard for mobile devices with over
1 billion installs in 2018. Gboard includes both typing features like text
autocorrection, next-word prediction and word completions as well as expression
features like emoji, GIFs and Stickers (curated, expressive illustrations and
animations). As both a mobile application and keyboard, Gboard has unique
constraints which lends itself well to both on-device inference and training.
First, as a keyboard application with access to much of what a user types into
their mobile device, Gboard must respect the user's privacy. Using FL allows us
to train machine learning models without collecting sensitive raw input from
users. Second, latency must be minimal; in a mobile typing environment, timely
suggestions are necessary in order to maintain relevance. On-device inference
and training through FL enable us to both minimize latency and maximize privacy.

In this paper, we use FL in a commercial, global-scale setting to train and
deploy a model to production for inference -- all without access to the
underlying user data. Our use case is search query suggestions~\cite{fedblog}:
when a user enters text, Gboard uses a \textit{baseline model} to determine and
possibly surface search suggestions relevant to the input. For instance, typing
\textit{``Let's eat at Charlie's''} may display a web query suggestion to search
for nearby restaurants of that name; other types of suggestions include GIFs and
Stickers. Here, we improve the feature by filtering query suggestions from the
baseline model with an additional \textit{triggering model} that is trained with
FL. By combining query suggestions with FL and on-device inference, we
demonstrate quality improvements to Gboard suggestions while enhancing user
privacy and respecting mobile constraints.

This is just one application of FL -- one where developers have never had access
to the training data. Other works have additionally explored federated
multi-task learning~\cite{fmtl}, parallel stochastic optimization~\cite{pso},
and threat actors in the context of FL~\cite{backdoor}. For next word
prediction, Gboard has also used FL to train a neural language model which
demonstrated better performance than a model trained with traditional
server-based collection and training~\cite{flkbd}. Language models have also
been trained with FL and differential privacy for further privacy
enhancements~\cite{differential-privacy, dplm}. By leveraging federated
learning, we continue to improve user experience in a privacy-advantaged manner.

This paper is organized as follows. In Section 2 we introduce FL, its
advantages and the enabling system infrastructure. Section 3 describes the
trained and deployed model architecture. Section 4 dives into our experience
training models with FL including training requirements and characteristics. In
Section 5, we deploy the federated trained model in a live inference experiment
and discuss the results, especially with respect to expected versus actual
metrics.

\section{Federated Learning and On-device Infrastructure}
\label{sec:federated_learning}

\subsection{Federated Learning} \label{sec:federated_learning_subsection}

The wealth of user interaction data on mobile devices, including typing,
gestures, video and audio capture, etc., holds the promise of enabling ever more
intelligent applications. FL enables development of such intelligent
applications while simplifying the task of building privacy into infrastructure
and training.

FL is an approach to distributed computation in which the data is kept at the
network edges and never collected centrally~\cite{fedlearn}. Instead, minimal,
focused model updates are transmitted, optionally employing additional
privacy-preserving technologies such as secure multiparty computation~\cite{secaggml}
and differential privacy~\cite{dplm, cpsgd, dlwdp}. Compared to traditional
approaches in which data is collected and stored in a central location, FL
offers increased privacy.

In summary, FL is best suited for tasks where one or more of the following hold:

\begin{enumerate}
  \item The task labels don't require human labelers but are naturally derived
  from user interaction.
  \item The training data is privacy sensitive.
  \item The training data is too large to be feasibly collected centrally.
\end{enumerate}

In particular, the best tasks for FL are those where (1) applies, and
additionally (2) and/or (3) apply.

\subsection{Privacy Advantages of Federated Learning}
\label{sec:privacy_advantages}

FL, specifically Federated Averaging~\cite{fedlearn}, in its most basic form
proceeds as follows. In a series of \textit{rounds}, the parameter server
selects some number of \textit{clients} to participate in training for that
round. Each selected client downloads a copy of the current model parameters
and performs some number of local model updates using its local training data;
for example, it may perform a single epoch of minibatch stochastic gradient
descent. Then the clients upload their model update -- that is, the difference
between the final parameters after training and the original parameters -- and
the server averages the contributions before accumulating them into the global
model.

In contrast to uploading the training data to the server, the FL approach has
clear privacy advantages even in its most basic form:

\begin{enumerate}
  \item Only the minimal information necessary for model training (the model
  parameter deltas) is transmitted. The updates will never contain more
  information than the data from which they derive, and typically will contain
  much less. In particular, this reduces the risk of deanonymization via joins
  with other data~\cite{sweeney}.
  \item The model update is ephemeral, lasting only long enough to be
  transmitted and incorporated into the global model. Thus while the model
  aggregator needs to be trusted enough to be given access to each client's
  model parameter deltas, only the final, trained model is supplied to end
  users for inference. Typically any one client's contribution to that final
  model is negligible.
\end{enumerate}

In addition to these advantages, FL can guarantee an even higher standard of
privacy by making use of two additional techniques. With \textit{secure
aggregation}~\cite{secaggml}, clients' updates are securely summed into a
single aggregate update without revealing any client's individual component
even to the server. This is accomplished by cryptographically simulating a
trusted third party. \textit{Differential privacy} techniques can be used in
which each client adds a carefully calibrated amount of noise to their update
to mask their contribution to the learned model~\cite{dplm}. However, since
neither of these techniques were employed in the present work, we will not
describe them in further detail here.

\subsection{System Description}
\label{sec:system_description}

\begin{figure*}
  \centering
  \includegraphics[width=\textwidth]{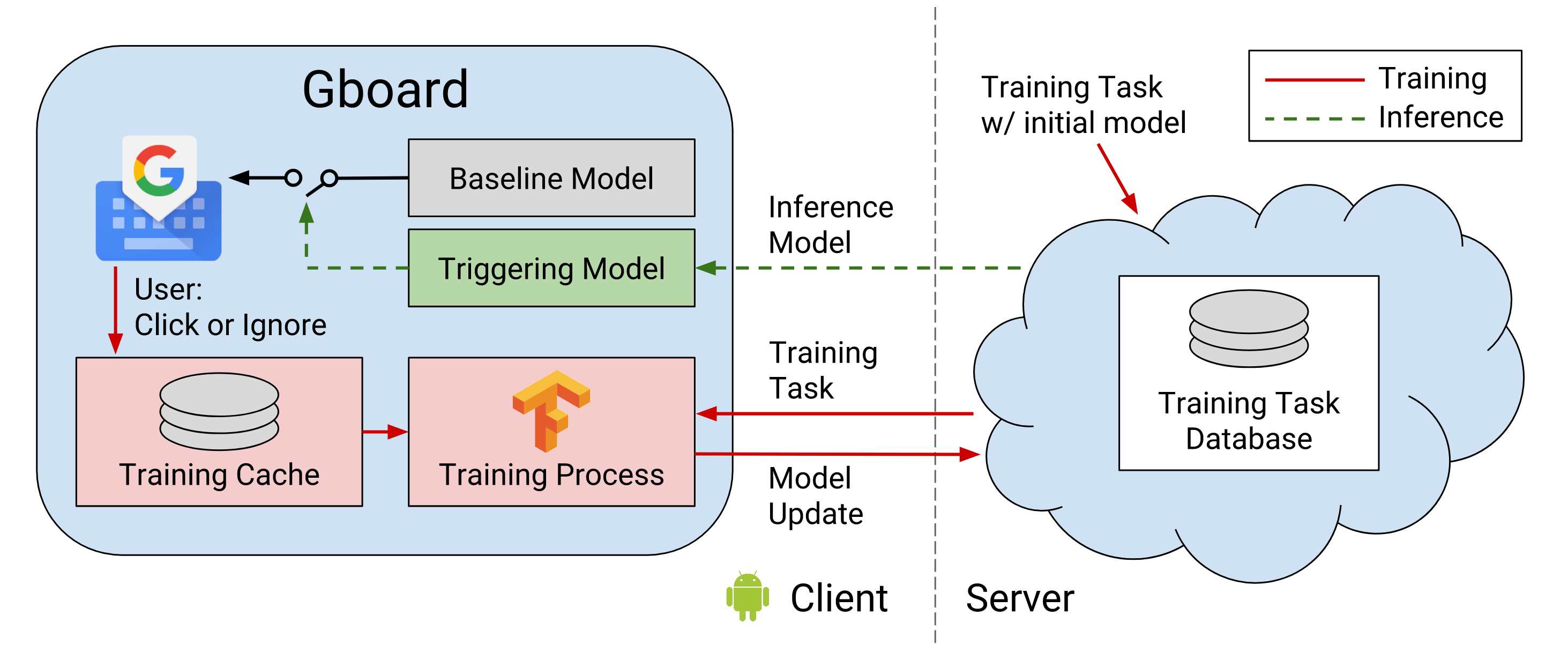}
  \caption{Architecture overview. Inference and training are on-device; model
  updates are sent to the server during training rounds and trained models
  are deployed manually to clients.}
  \label{fig:system_arch}
\end{figure*}

In this section we provide a brief technical description of the client and
server side runtime that enables FL in Gboard by walking through the process of
performing training, evaluation and inference of the query suggestion triggering
model.

As described earlier, our use case is to train a model that predicts whether
query suggestions are useful, in order to filter out less relevant queries. We
collect training data for this model by observing user interactions with the
app: when surfacing a query suggestion to a user, a tuple \textit{(features;
label)} is stored in an on-device \textit{training cache}, a SQLite based
database with a time-to-live based data retention policy.

\begin{itemize}
  \item \textit{features} is a collection of query and context related
  information
  \item \textit{label} is the associated user action from \textit{\{clicked,
  ignored\}}.
\end{itemize}

This data is then used for on-device training and evaluation of models provided
by our servers.

A key requirement for our on-device machine learning infrastructure is to have
no impact on user experience and mobile data usage. We achieve this by using
Android's JobScheduler to schedule background jobs that run in a separate Unix
process when the device is idle, charging, and connected to an unmetered
network, and interrupt the task when these conditions change.

When conditions allow -- typically at night time when a phone is charging and
connected to a Wi-Fi network -- the client runtime is started and checks in with
our server infrastructure, providing a \textit{population name} identifier, but
no information that could be used to identify the device or user. The server
runtime waits until a predefined number of clients for this population have
connected, then provides each with a \textit{training task} that contains:

\begin{itemize}
  \item a model consisting of a TensorFlow graph and
  checkpoint~\cite{tensorflow}
  \item metadata about how to execute the model (input + output node names,
  types and shapes; operational metrics to report to the server such as loss,
  statistics of the data processed). Execution can refer, but is not limited
  to, training or evaluation passes.
  \item selection criteria used to query the training cache (e.g. filter data
  by date)
\end{itemize}

The client executes the task using a custom task interpreter based on
TensorFlow Mobile~\cite{tflite}, a stripped down Android build of the
TensorFlow runtime. In the case of training, a task-defined number of epochs
(stochastic gradient descent passes over the training data) are performed, and
the resulting updates to the model and operational metrics are anonymously
uploaded to the server. There -- again using TensorFlow -- these ephemeral
updates are aggregated using the Federated Averaging algorithm to produce a new
model, and the aggregate metrics allow for monitoring training progress.

To balance the load across more devices and avoid over- or under-representing
individual devices in the training procedure, the client receives a minimum
delay it should wait before checking in with the server again. In parallel to
these on-device training tasks that modify the model, we also execute on-device
evaluation tasks of the most recent model iteration similar to data center
training to monitor progress, and estimate click-threshold satisfying
requirements such as retained impression rate. Upon convergence, a trained
checkpoint is used to create and deploy a model to clients for inference.

\section{Model Architecture}
\label{sec:model}

\begin{figure*}
  \centering
  \includegraphics[width=\textwidth]{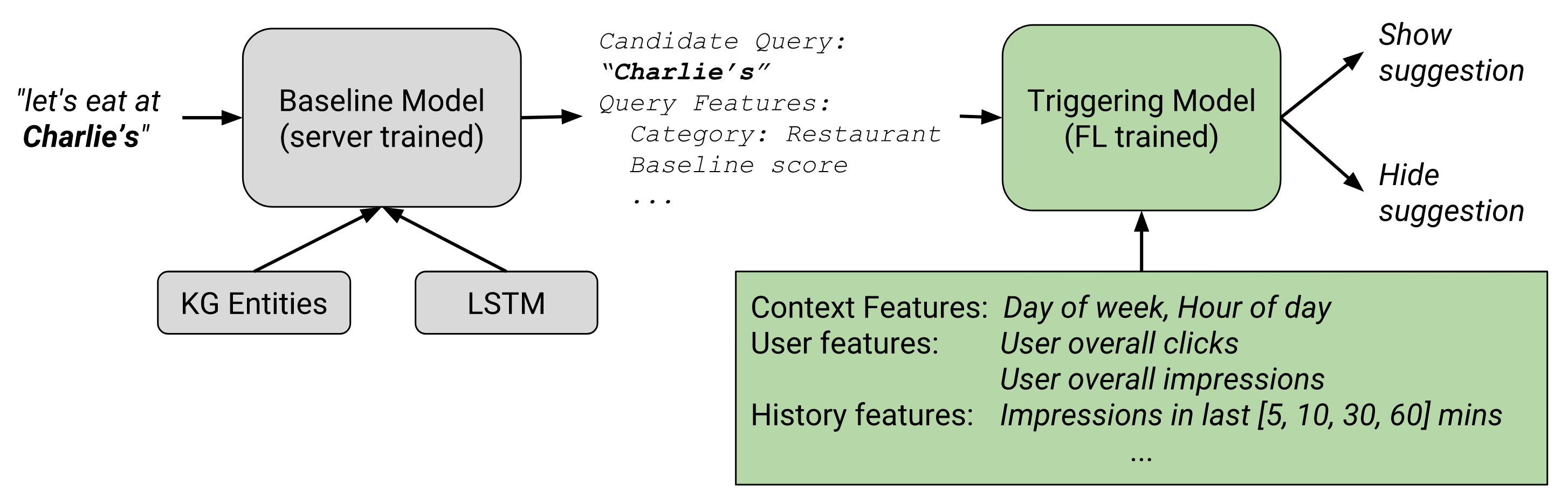}
  \caption{Setup of the baseline model (traditionally server trained) with the
  triggering model (federated trained). The baseline model generates candidates
  and the triggering model decides whether to show the candidate.}
  \label{fig:model_arch}
\end{figure*}

In this section, we describe the model setup used to improve Gboard's query
suggestion system. The system works in two stages -- a traditionally server-side
trained \textit{baseline model} which generates query candidates, and a
\textit{triggering model} trained via FL (Figure~\ref{fig:model_arch}). The goal
is to improve query click-through-rate (CTR) by taking suggestions from the
baseline model and removing low quality suggestions through the triggering
model.

\subsection{Baseline Model} \label{sec:baseline_model}

First, we use a \textit{baseline model} for query suggestion, trained offline
with traditional server-based machine learning techniques. This model first
generates query suggestion candidates by matching the user's input to an
on-device subset of the Google Knowledge Graph (KG)~\cite{kg}.

It then scores these suggestions using a Long Short-Term Memory
(LSTM)~\cite{lstm} network trained on an offline corpus of chat data to detect
potential query candidates. This LSTM is trained to predict the KG category of a
word in a sentence and returns higher scores when the KG category of the query
candidate matches the expected category. With on-device training, we expect to
improve over the baseline model by making use of user clicks and interactions --
signals which are available on-device for federated training.

The highest scoring candidate from the baseline model is selected and displayed
as a query suggestion (an impression). The user then either clicks on or
ignores the suggestion. We store these suggestions and user interactions in the
on-device training cache, along with other features like time of day, to
generate training examples for use in FL.

\subsection{Triggering Model} \label{sec:triggering_model}

The task of the federated trained model is designed to take in the suggested
query candidate from the baseline model, and determine if the suggestion should
or should not be shown to the user. We refer to this model as the
\textit{triggering model}. The triggering model used in our experiments is a
logistic regression model trained to predict the probability of a click; the
output is a score for a given query, with higher scores meaning greater
confidence in the suggestion.

When deploying the triggering model, we select a threshold $\tau$ in order to
reach a desired triggering rate, where a higher threshold is stricter and
reduces the trigger rate of suggestions. Tuning the triggering threshold allows
us to balance the tradeoff between providing value to the user and potentially
hiding a valuable suggestion. As a logistic regression model, the score is in
logit space, where the predicted probability of a click is the logistic sigmoid
function applied to the score. In model training and deployment, we evaluated
performance at uniformly spaced thresholds.

A feature-based representation of the query is supplied to the logistic
regression model. Below are some of the features we incorporate in our model:

\textbf{Past Clicks and Impressions} The number of impressions the current user
has seen on past rich suggestions, as well as the number of times they have
clicked on a suggestion, both represented as log transformed real values. The
model takes in both overall clicks and impressions, as well as clicks and
impressions broken down by KG category. This allows the model to personalize
triggering based on past user behavior.

\textbf{Baseline Score} The score output by the baseline model, represented as
a binned, real-valued feature. This score is derived from an LSTM model over
the input text, which incorporates context into the model.

\textbf{Day of Week, Hour of Day} These temporal features, represented as
one-hot vectors, allow the model to capture temporal patterns in query
suggestion click behavior.

Using a logistic regression model as an initial use case of FL has the advantage
of being easily trainable given the convexity of the error function, as compared
to multi-layer neural networks. For our initial training, we have a limited
number of clients and training examples, which makes it impractical to train
models with a large number of parameters. Furthermore, the label is binary and
heavily skewed (we have many more impressions than clicks). However a benefit of
logistic regression is that it is possible to interpret and validate the
resulting trained model by directly inspecting the model weights. In other
environments with more data or clients, more complex neural network models can
be trained with FL~\cite{flkbd}.

\section{Training with Federated Learning}
\label{sec:training}

Here we describe the conditions we require for FL, as well as observations from
training with FL.

\subsection{Federated Training Requirements} \label{sec:training_reqs}

For training with FL, we enforce several constraints on both devices and FL
tasks.

\bigskip

To participate in training, a client device must meet the following
requirements:

\textbf{Environmental Conditions} Device must be charging, on unmetered network
(typically Wi-Fi), and idle. This primarily translates to devices being charged
overnight and minimizes impact to the user experience.

\textbf{Device Specifications} Device must have at least 2GB of memory and
operate Android SDK level 21+.

\textbf{Language Restriction} Limited to en-US and en-CA users. However, many of
our training clients are in India and other countries which commonly default to
an Android locale of en-US, causing unexpected training populations. While this
is fixed in later infrastructure iterations, the work presented here does have
this skew.

\bigskip

During federated training, we apply the following server constraints to our FL
tasks:

\textbf{Goal Client Count} The target number of clients for a round of
federated training, here 100.

\textbf{Minimum Client Count} The minimum number of clients required to run a
round. Here 80, i.e. although we ideally want 100 training clients, we will run
a round even if we only have 80 clients.

\textbf{Training Period} How frequently we would like to run rounds of
training. Here 5 min.

\textbf{Report Window} The maximum time to wait for clients to report back with
model updates, here 2 minutes.

\textbf{Minimum Reporting Fraction} The fraction of clients, relative to the
actual number of clients gathered for a round, which have to report back to
commit a round by the end of the Report Window. Here 0.8.

\subsection{Federated Training} \label{sec:federated_training}

\begin{figure*}
  \centering
  \includegraphics[width=\textwidth]{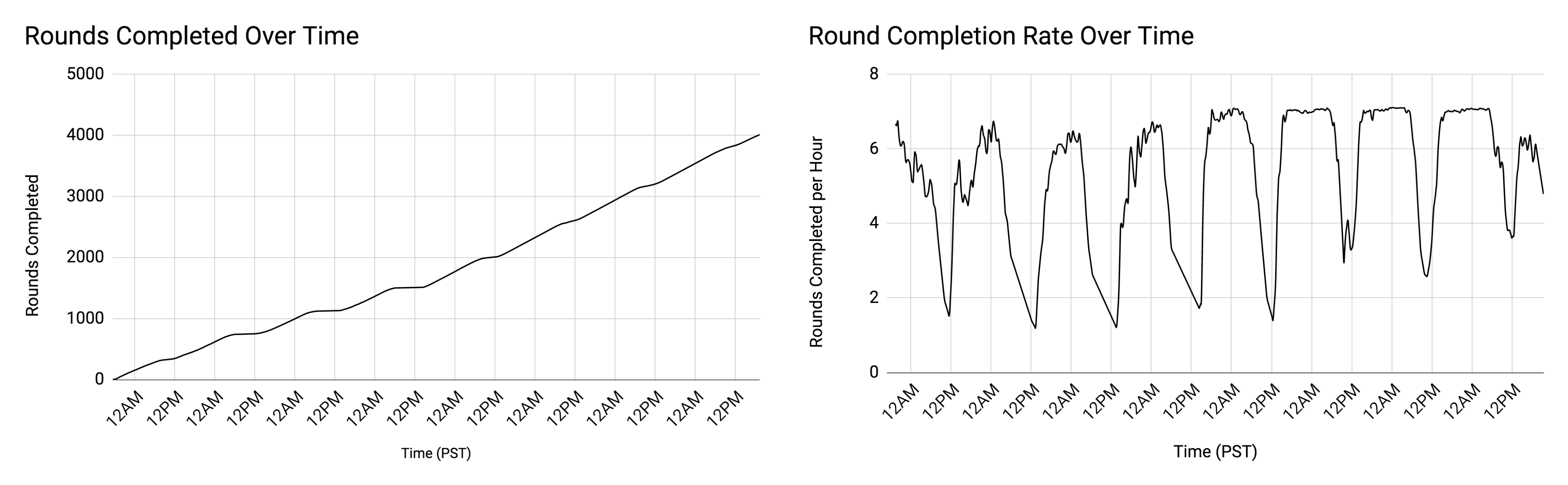}
  \caption{Round completion over time and round completion rate over time,
  times are in PST. Rounds progress faster at night when more devices are
  charging and on an unmetered network.}
  \label{fig:rounds_completed_combined}
\end{figure*}

\begin{figure}[h]
  \centering
  \includegraphics[width=1\columnwidth]{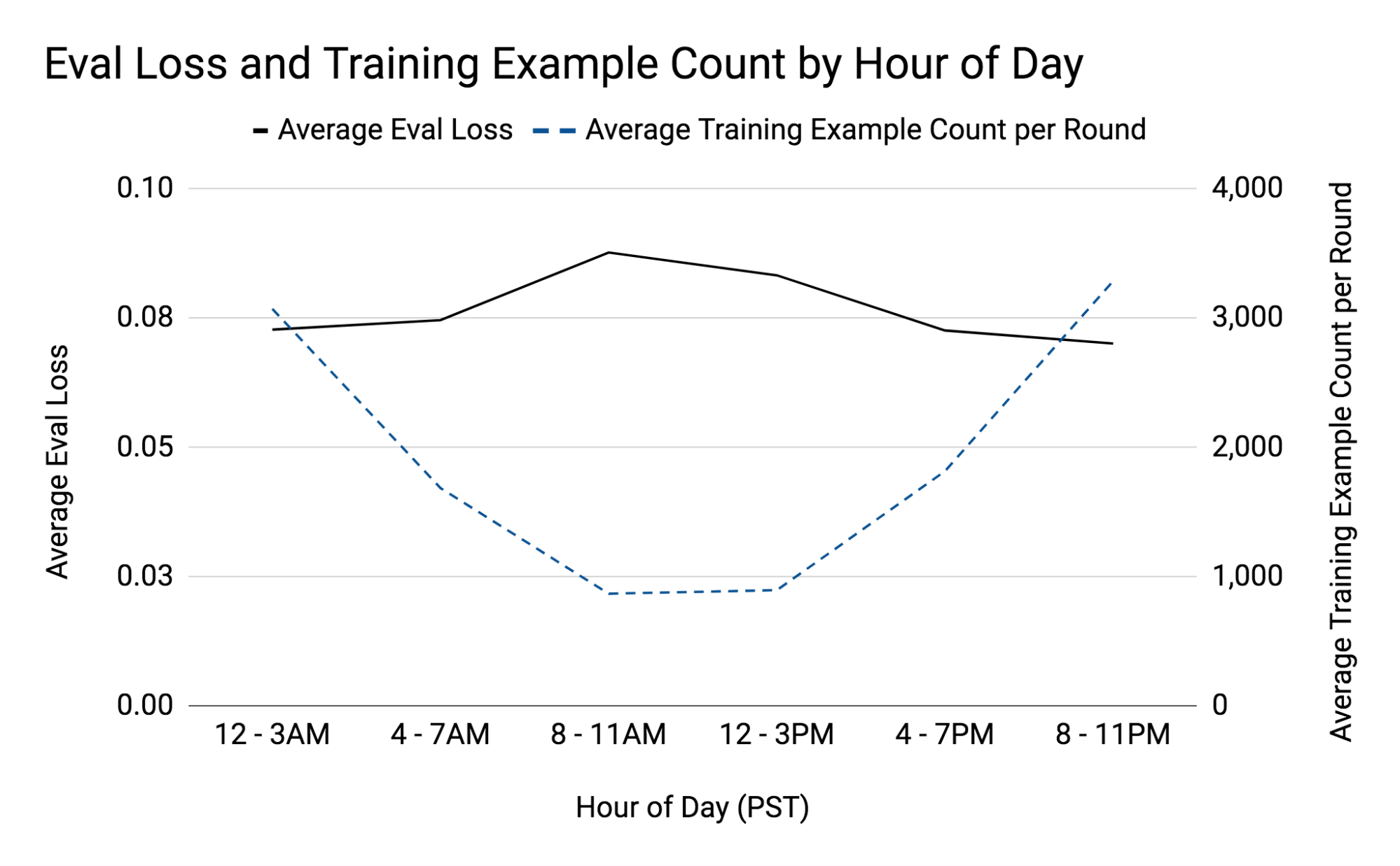}
  \caption{Eval loss and training example count over time, times are in PST,
  hour ranges inclusive. Training example count is highest in the evening as
  more devices are available. In contrast, eval loss is highest during the
  day when few devices are available and those available represent a skewed
  population.}
  \label{fig:eval_loss_and_example_count}
\end{figure}

\begin{figure}[h]
  \centering
  \includegraphics[width=1\columnwidth]{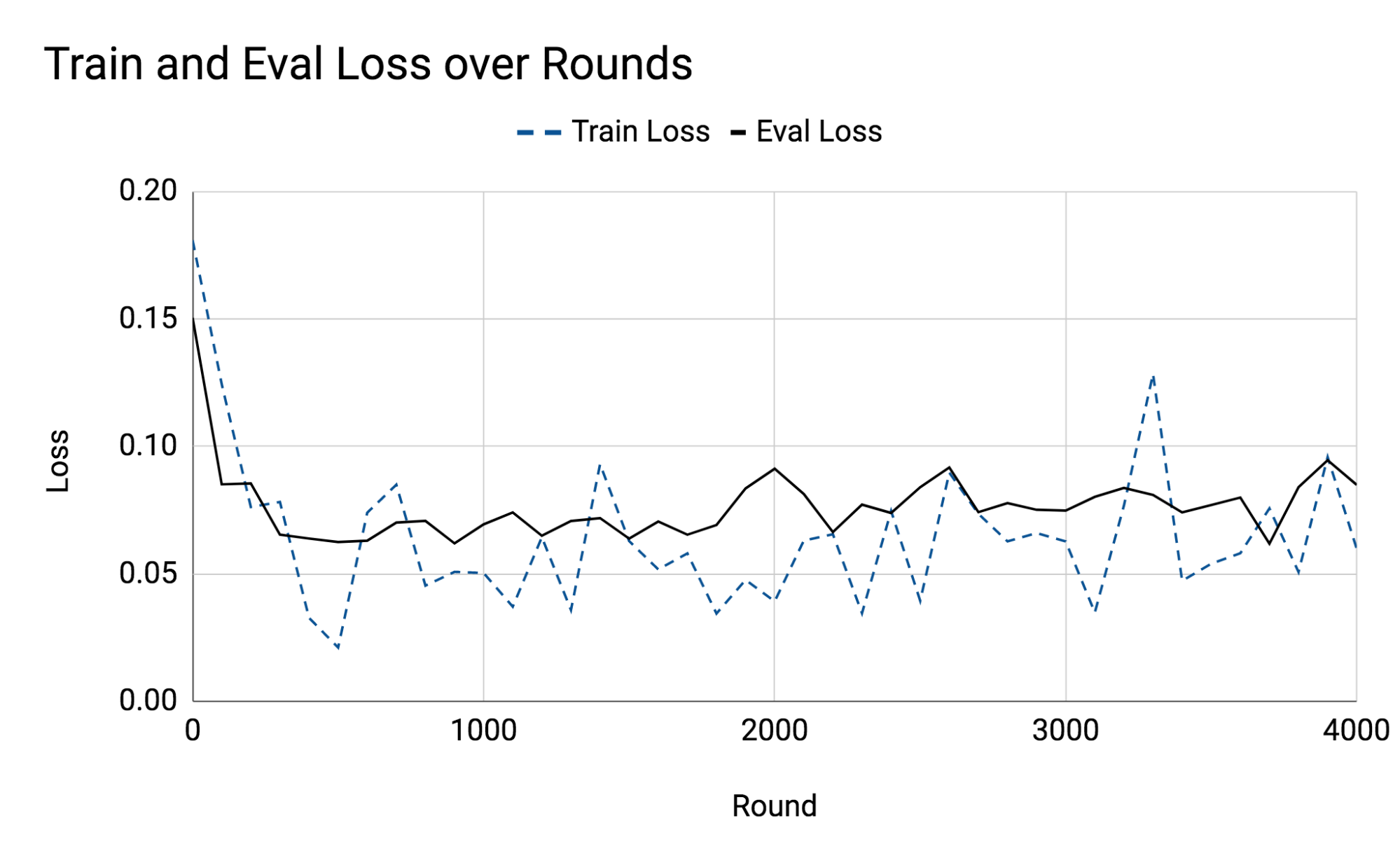}
  \caption{Train and eval loss of the logistic regression triggering model
  over rounds (bucketed to 100 rounds).}
  \label{fig:train_and_eval_loss}
\end{figure}

\begin{figure}[h]
  \centering
  \includegraphics[width=1\columnwidth]{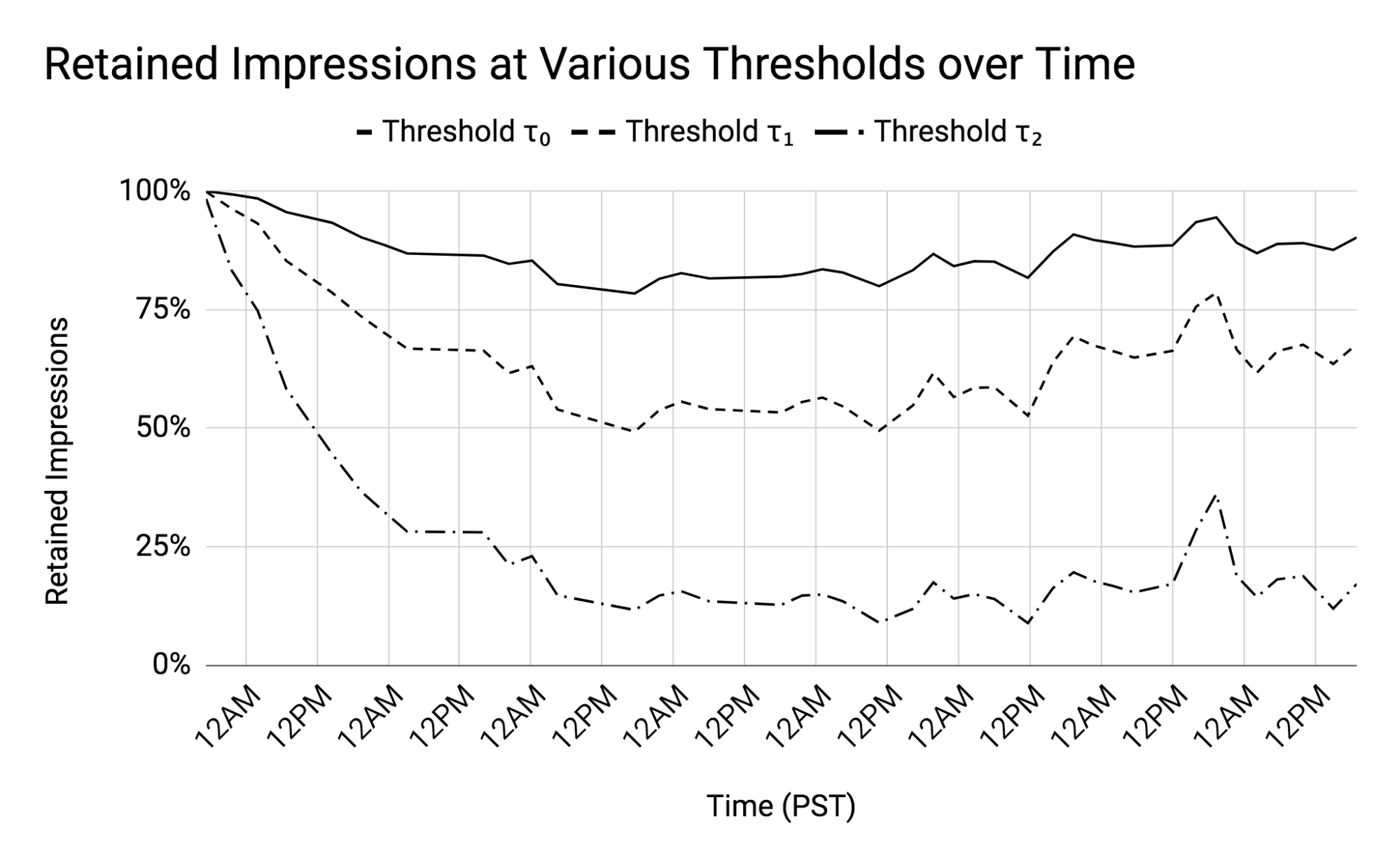}
  \caption{Retained impressions at thresholds $\tau_0 < \tau_1 < \tau_2$,
  uniformly spaced, over time.}
  \label{fig:retained_impressions}
\end{figure}

From our training of the Triggering Model with FL, we make several
observations.

Since we only train on-device when the device is charging, connected to an
unmetered network, and idle, most training occurs in the evening hours, leading
to diurnal patterns. As a result, training rounds progress much more quickly at
night than during the day as more clients are available, where night/day are in
North America time zones since we are targeting devices with country United
States and Canada.

Figure~\ref{fig:rounds_completed_combined} shows training progress across a
week (times in PST) while Figure~\ref{fig:eval_loss_and_example_count} shows
our model training and eval loss bucketed by time of day. Note that most round
progression is centered around midnight, while round completion around noon is
nearly flat. While there are still clients available during non-peak hours,
fewer rounds are completed due to fewer available clients as well as client
contention from running other training tasks, e.g. to train language
models~\cite{flkbd}. As a result, it is more difficult to satisfy the training
task parameters -- in our case 100 clients with 80\% reporting back within two
minutes for a successful round.

Furthermore, devices which are training during off-peak hours (during the day
in North America) have an inherent skew. For example, devices which require
charging during the day may indicate that the phone is an older device with a
battery which requires more frequent charging.

Similarly, our country restriction was based on the device's Android country so
any device with a locale set to en-US or en-CA would participate in training
even if the user was not geographically located in the US or Canada. In
particular, many devices in and around India are set by default to Android
locale en-US and are not modified by end-users; as a result, many of our devices
training during the day (PST) are devices located in and around India (where it
is evening). These developing markets tend to have less reliable network and
power access which contributes to low round completions~\cite{internet,
unenergy, indiaqos}.

The diurnal distribution of user populations also contributes to regular cycles
in model metrics, including training and eval loss, during federated training.
Since the devices training on clients during the day are often unexpected user
populations with bias from the main population, the overall model loss tends to
increase during the day and decrease at night. Both of these diurnal effects are
evident in Figure~\ref{fig:train_and_eval_loss} which depicts the average eval
loss and the average training examples per round, bucketed by time of day. Note
that training example count is highest in the evening as more devices are
available. In contrast, eval loss is highest during the day when few devices are
available and those available represent a skewed population.

In addition to the loss, we can measure other metrics to track the performance
of our model during training. In Figure~\ref{fig:retained_impressions}, we plot
the retained impressions at various thresholds over time.

As a whole, training with FL introduces a number of interesting diurnal
characteristics. We expect that as development of FL continues, overall
training speed of FL will increase; however, the nature of globally-distributed
training examples and clients will continue to be inherent challenges with FL.

\subsection{Model Debugging Without Training Example Access}
\label{sec:model_debugging}

Typically a trained machine learning model can be validated by evaluating its
performance on a held-out validation set. The FL analogue to such a central
validation set is to perform evaluation by pushing the trained model to devices
to score it on their own data and report back the results. However, on-device
metrics like loss, CTR and retained impressions do not necessarily tell the
whole story that traditionally inspecting key training examples might provide.
For this reason it is valuable to have tools for debugging models without
reference to training data.

During model development, synthetically generated proxy data was used to
validate the model architecture to select ballparks for basic hyperparameters
like learning rate. The generated data encoded some basic first-order statistics
like aggregate click-through-rate, an expectation that some users would click
more than others or at different times of day, etc. The model was also validated
with integration and end-to-end testing on a handful of realistic hand --
constructed and donated examples. The combination of synthetic and donated data
enabled model development, validated that the approach learned patterns we'd
encoded into the data, and built confidence that it would learn as expected with
data generated by the application.

To validate the model after training, we interpreted the coefficients of our
logistic regression model via direct examination of the weights in order to gain
insight into what the model had learned. We determined that FL had produced a
reasonable model (reasonable enough to warrant pushing to devices for live
inference experiments) considering that:

\begin{itemize}
  \item the weights corresponding to query categories had intuitive values
  \item the weights corresponding to binned real-valued features tended to have
  smooth, monotone progressions
  \item the weights of more common features were larger in absolute value, i.e.
  the model came to rely on them more due to their frequency.
\end{itemize}

Manual inspection of the weights also uncovered an unusual pattern that
revealed a way to improve future model iterations. One binned real-valued
feature had zero weight for most of its range, indicating that the expected
range of the feature was too large. We improved future iterations of the model
by restricting the feature to the correct range so the binned values (which did
not change in number) gave more precision within the range. This is just one
example approach to the broader domain of debugging without training example
access.

\section{Live Results and Observations}
\label{sec:results}

\begin{table*}
  \centering
  \begin{tabular}{ccccccc} \toprule
    & \multicolumn{3}{c}{Training Metrics From Federated Training} &
    \multicolumn{3}{c}{Live Metrics From Live Experiments}     \\
    Threshold & $\Delta$CTR               & Retained impressions  &
    Retained
    clicks & $\Delta$CTR              & Retained impressions & Retained
    clicks \\ \hline \midrule
    $\tau_0$        & +3.01\%            & 93.44\%               &
    96.25\%         & N/A               & N/A                  &
    N/A         \\
    $\tau_1$        & +17.40\%  & 75.60\%               &
    88.76\%         & +14.52\% & 67.33\%              &
    77.11\%         \\
    $\tau_2$        & \textbf{+42.19\%}  & 28.45\%               &
    40.45\%         & \textbf{+33.95\%} & 24.18\%              &
    32.39\%         \\
    \bottomrule
  \end{tabular}
  \caption{Metrics during training and live model deployment, where $\tau_0 <
  \tau_1 < \tau_2$, uniformly spaced.  A greater threshold indicates a
  stricter quality bar for suggestions. }
    \label{tab:metrics}
\end{table*}

\begin{table}
  \centering
  \begin{tabular}{ccc} \toprule
    Model             & Live $\Delta$CTR & Live Retained Clicks \\ \hline
    \midrule
    Model Iteration 1 & +14.52\%  & 77.11\%               \\
    Model Iteration 2 & +25.56\%  & 63.39\%               \\
    Model Iteration 3 & \textbf{+51.49\%}  &
    \textbf{82.01\%}            \\
    \bottomrule
  \end{tabular}
  \caption{Change in CTR over several trained and deployed models. In later
  development we trained an LSTM model which performed the best in terms of
  $\Delta$CTR and Retained Clicks.}
  \label{tab:metrics_iterated}
\end{table}

After training and sanity checking our logistic regression model, we deployed
the model to live users. A model checkpoint from the server was used to build
and deploy an on-device inference model that uses the same featurization flow
which originally logged training examples on-device. The model outputs a score,
which is then compared to a threshold to determine if the suggestion is shown
or hidden. By selecting various thresholds, we experiment with different
operating points, trading off CTR for retained impressions and clicks. We
deployed our models on a population with the same locale restrictions as our
training population (en-US and en-CA).

Comparing our expected $\Delta$CTR in training metrics to our actual
$\Delta$CTR in live experiments, our live deployments reflect a successful
improvement in CTR (Table~\ref{tab:metrics}). However, while we capture a
majority of the expected improvements, we do observe a slight drop between
expected and actual $\Delta$CTR.

Some hypotheses for this $\Delta$CTR follow:

\textbf{Environmental Conditions} Since we require devices to be charging and on
unmetered networks, this biases our training towards devices and users with
these conditions available. In particular, many users in developing countries do
not have reliable access to either stable power or stable unmetered
networks~\cite{unenergy, internet}.

\textbf{Device Specifications} We restricted device training to devices with 2GB
of RAM, however our deployment was on all devices without a minimum RAM
requirement. This causes skew in the training vs deployment population in terms
of device specifications.

\textbf{Successful Training Clients} Recall that our federated training
configuration only requires 80\% of selected devices to respond in order to
close a round. This skews our training population towards higher end devices
with more stable networks, as lower end devices are more unstable from a device
and network perspective.

\textbf{Evaluation and Training Client Overlap} With our federated training
infrastructure, the client selected for training and eval rounds are not
necessarily mutually exclusive. However, given that training and eval rounds
only select a small subset of the overall training population, we expect the
overlap to be $<$0.1\% and have minimal impact on performance skew.

In addition to the above hypotheses, more general sources of skew may also
apply, such as model drift due to training and deploy time offsets. As FL
continues to mature, we expect that the delta between expected and actual
metrics will narrow over time.

The results detailed here were only the first in a sequence of models trained,
evaluated, and launched with FL. Successive iterations differed in that they
were trained longer on more users' data, had better tuned hyperparameters, and
incorporated additional features. The results of these successive iterations are
shown in Table~\ref{tab:metrics_iterated}, but we do not describe in depth here
all the changes made between these iterations. One noteworthy addition in the
final model was the inclusion of an LSTM-based featurization of the typed text,
which was co-trained with the rest of the logistic regression model, allowing
the model to better understand the typed context. Our iterations with FL have
demonstrated the ability to develop effective models in a privacy advantaged
manner without direct access to the underlying user data.

\section{Conclusion}
\label{sec:conclusion}

In this work, we applied FL to train, evaluate and deploy a logistic regression
model without access to underlying user data to improve keyboard search
suggestion quality. We discussed observations about federated learning including
the cyclic nature of training, model iteration without direct access to training
data, and sources of skew between federated training and live deployments. This
training and deployment is one of the first end-to-end examples of FL in a
production environment, and explores a path where FL can be used to improve user
experience in a privacy-advantaged manner.

\section{Acknowledgments}
\label{sec:acknowledgments}

The authors would like to thank colleagues on the Gboard and Google AI team for
providing the federated learning framework and for many helpful discussions.

\bibliographystyle{IEEEbib}
\bibliography{paper}

\end{document}